\documentclass[journal]{IEEEtran} 

\usepackage[utf8]{inputenc}         
\usepackage[T1]{fontenc}            
\usepackage{hyperref}               
\usepackage{url}                    
\usepackage{nicefrac}               
\usepackage{microtype}              
\usepackage{amsfonts}               
\usepackage{amsmath}                
\usepackage[dvipsnames]{xcolor}     
\usepackage{float}                  
\usepackage{graphicx}               
\graphicspath{{images/}}
\DeclareGraphicsExtensions{.pdf,.PDF,.jpg,.JPG,.jpeg,.JPEG,.png,.PNG}
\usepackage{booktabs}               
\usepackage{multicol}               
\usepackage{multirow}               
\usepackage{booktabs}               
\usepackage{array}                  
\newcolumntype{C}[1]{>{\centering\arraybackslash}p{#1}}

\usepackage[nolist]{acronym}        
\usepackage{verbatim}               
\usepackage{comment}                

\usepackage{textcomp}
\usepackage[numbers]{natbib}
\usepackage{makecell}
\usepackage[super]{nth} 
\usepackage{amssymb}    
\usepackage{graphicx}   
\usepackage{subcaption} 
\usepackage{arydshln}   

\newcommand{\ie}{\textit{i}.\textit{e}.,}

\usepackage{xcolor}

\newcommand\tocite[1]{\textcolor{blue}{[REFERENCE]}}

\begin{document}

\title{Segmentation and Smoothing Affect Explanation Quality More Than the Choice of Perturbation-based XAI Method for Image Explanations}


\author{
    \IEEEauthorblockN{
        \textbf{Gustav~Grund~Pihlgren}\IEEEauthorrefmark{1} \and
        \textbf{Kary~Främling}\IEEEauthorrefmark{1}
    }\\
    \vspace{0.2cm}

    \IEEEauthorblockA{
        \IEEEauthorrefmark{1}%
        \textit{Dept. of Computing Science} \\
        Umeå University, Sweden\\
    }
}

\markboth{}{First author et al. : Title}

\maketitle

\thispagestyle{empty}

\begin{abstract}

Perturbation-based post-hoc image explanation methods are commonly used to explain image prediction models.
These methods perturb parts of the input to measure how those parts affect the output.
Since the methods only require the input and output, they can be applied to any model, making them a popular choice to explain black-box models.
While many different methods exist and have been compared with one another, it remains poorly understood which parameters of the different methods are responsible for their varying performance.



This work uses the Randomized Input Sampling for Explanations (RISE) method as a baseline to evaluate many combinations of mask sampling, segmentation techniques, smoothing, attribution calculation, and per-segment or per-pixel attribution, using a proxy metric.
The results show that attribution calculation, which is frequently the focus of other works, has little impact on the results.
Conversely, segmentation and per-pixel attribution, rarely examined parameters, have a significant impact.

The implementation of and data gathered in this work are available online~\footnote{\url{https://github.com/guspih/post-hoc-image-perturbation}}\footnote{\url{https://bit.ly/smooth-mask-perturbation}}.

\end{abstract}

\section{Introduction}
\label{toc:introduction}







Over the past decade, deep neural networks (DNN) have proven proficient at solving computer vision tasks~\cite{sarraf2021comprehensive}.
However, the black-box nature of DNNs causes issues, including difficulties in understanding when the model is wrong, a lack of trust in the models, and legal issues~\cite{barredo2020explainable}.
The goal of the field of Explainable Artificial Intelligence (XAI) is to make AI models more transparent to mitigate these issues.

The complexity of DNNs makes it difficult to explain the model in its entirety.
Instead, methods for explaining DNNs tend to focus on explaining individual predictions made by the model~\cite{linardatos2021explainable}.
Some research focuses on developing inherently interpretable models~\cite{schwalbe2023comprehensive}.
However, these methods are bound to specific models and cannot be applied to the vast amount of non-interpretable models that exist and are being developed~\cite{linardatos2021explainable}.
In these cases, it is common to use post-hoc explanations that are separate from the models and can be applied during prediction~\cite{madsen2022posthoc}.

Many different types of post-hoc XAI methods exist to explain computer vision models.
However, many methods can only be used on models that satisfy certain criteria, such as differentiability in the case of gradient-based methods.
This work instead concerns a type of post-hoc explanation method that does not place any restriction on the models being used; so-called perturbation-based methods.
Perturbation-based explanations work by analyzing how the model's predictions change as the original input is perturbed.
For image prediction, this is commonly used to find the regions of the image that cause the model to make a certain prediction.
This can be used to assess if a specific model or prediction is sound and help locate important features for a user.

Since the information in images is generally found in the relationships between many pixels~\cite{pihlgren2023perceptual}, perturbing individual pixels is unlikely to have much impact on the prediction.
As such, perturbations are typically made on larger pixel segments.
Depending on the method, these segments are either perturbed one at a time or several at once, with different sampling methods for determining which segments to perturb.

\begin{figure}[t] 
    \centering
    \includegraphics[width=\columnwidth]{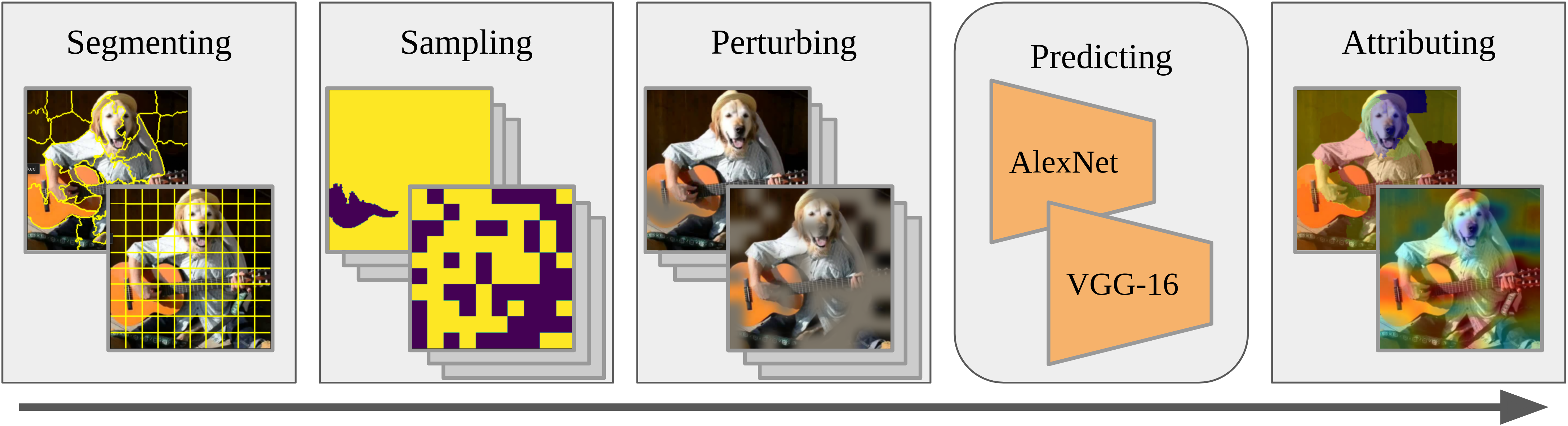}
    \caption{The pipeline for perturbation-based image attribution used in this work. The image is segmented, samples indicating what segments to perturb are drawn, the sampled segments are perturbed, the model to explain makes predictions for the perturbed samples, and the input-output pairs are used to calculate per-segment and per-pixel attribution.}
    \label{fig:perturbation_pipeline}
\end{figure}

The general pipeline for calculating perturbation-based image explanations consists of segmenting, sampling, perturbing, predicting, and attributing, as shown in Fig.~\ref{fig:perturbation_pipeline}.
The image is split into segments, and a number of samples are drawn, indicating which segments should be perturbed.
For each sample, a new image is created by perturbing the indicated segments in some way.
Perturbation often consists of occluding the segments with a solid color~\cite{ivanovs2021perturbation}, but other distortions such as inpainting have also been used~\cite{blucher2024decoupling}.
The model output from these perturbed inputs can then be used to attribute influence to the segments based on how the output changes when they are perturbed or not.
There are many ways to calculate attribution based on the input-output pairs, such as average output when a segment is included~\cite{petsiuk2018rise} or excluded~\cite{robnik2008explaining}.
Another method is to train a surrogate model to predict the output based on the perturbations and use the learned parameters as attribution~\cite{ribeiro2016should, lundberg2017unified}.

The influence attributed to each segment is often visualized using attribution maps.
Attribution maps are heat maps showing high influence using one color and fading to another color to show low influence.
By analyzing the attribution map, a human can understand the model's behavior.
For example, a prediction with a high influence on a non-salient part of the image can indicate problems with the model.
Using attribution map explanations increases the joint human-AI performance on some tasks~\cite{nguyen2021effectiveness}.

Since influence is attributed based on which segments are perturbed, most perturbation-based methods assign attribution to entire segments, but cannot differentiate the influence between pixels within a segment.
The Randomized Input Sampling for Explanations (RISE) method solves this by using smooth perturbations, where pixels closer to the segment center are perturbed more~\cite{petsiuk2018rise}.
With this approach, a per-pixel attribution can be calculated by assigning each pixel the influence of each segment, weighted by how perturbed the pixel was for that segment.

As for most papers on perturbation-based explanations for images, in~\cite{petsiuk2018rise}, RISE is described as an entire pipeline from segmenting to pixel attribution.
As such, it is not clear which parts of the RISE method, such as per-pixel attribution, that is responsible for its performance.
This work explores the individual parameters of the perturbation-based pipeline to evaluate which parameters have the most impact on performance.
This is done by taking a version of the original RISE pipeline and evaluating the original parameters alongside a variety of additional segmentation, sampling, perturbation, and attribution methods.

Due to the large number of parameter combinations to evaluate, human evaluation is not feasible for this study.
Instead, this work uses a faithfulness metric dubbed Symmetric Relevance Gain (SRG)~\cite{blucher2024decoupling}.
The evaluations are carried out on the ImageNet validation set~\cite{deng2009imagenet} for three different CNNs~\cite{krizhevsky2012imagenet, simonyan2015very, he2016deep} using both per-segment and per-pixel attributions.
The different pipeline parameters that have been combined and evaluated are shown in Table~\ref{tab:pipeline_parameters}.

The results show that the method of calculating the attribution, which is typically what is highlighted as the most important part, has little impact on performance.
Instead, segmentation and sampling, which are often overlooked, have a significant impact on performance.
This calls into question studies that compare the original implementations of attribution calculation methods without correcting for the other parameters of those implementations.
Additionally, per-pixel attribution gives a significantly higher average performance compared to per-segment attribution.

\subsection{Contributions}
The primary contributions of this work are as follows.
\begin{itemize}
    \item Evaluating the different parameters of the perturbation-based image attribution pipeline, finding that all parameters can have a significant impact on performance.
    \item Adapting the smooth-mask enabled per-pixel attribution used in RISE to work with other perturbation-based attribution methods.
    \item Improving the RISE pipeline by using Gaussian filtering instead of bilinear upsampling, which enables the use of many other segmenting methods.
    \item Generalizing the Probabilistic Difference Analysis (PDA)~\cite{robnik2008explaining} method to work with perturbing multiple segments per sample.
\end{itemize}

\begin{table}[t] 
    \centering
    \caption{The different parameters of each step of the perturbation-based image explanation pipeline evaluated in this work. All possible combinations of these parameters were evaluated.}
    \label{tab:pipeline_parameters}
    \resizebox{\linewidth}{!}{%
    \begin{tabular}{l l l l l}
    \toprule
        \makecell[l]{Segmenting +\\Perturbing} & Sampling & Samples & Model & Attribution\\\hline
        Grid+Bilinear & Only one & 4000/8000 & AlexNet & CIU\\
        Grid+Gaussian & All but one & 400 & VGG-16 & PDA\\
        SLIC+Gaussian & Random & 50 & ResNet & LIME\\
                      & Entropic & & & SHAP\\
                      & & & & RISE\\
    \bottomrule
    \end{tabular}
    }
\end{table}





\section{Methodology}
\label{toc:methodology}


This work evaluates different segmenting, sampling, perturbing, and attribution methods using three different sample sizes as listed in Table~\ref{tab:pipeline_parameters}.
All possible pipelines combining one of each parameter are evaluated by explaining the predictions made by three different ImageNet~\cite{deng2009imagenet} pretrained CNNs as measured by the SRG metric~\cite{blucher2024decoupling}.
The different parts of the experiments are described in detail in the following subsections.


\subsection{Segmenting}
This work evaluates two segmenting techniques: Grid and SLIC~\cite{achanta2012slic} segmentation.
Grid segmentation splits the image a given number of times horizontally and vertically.
SLIC is a rule-based algorithm that iteratively calculates segment "centers", assigns each pixel to the closest center in a color-position space, and recalculates the segment centers repeatedly until convergence.

The experiments use the same $7\times7$ segmentation grid as the original RISE implementation~\cite{petsiuk2018rise}.
To make the SLIC segmentation as similar to the grid implementation as possible, SLIC is instantiated with 49 segment centers in the experiments.
The default scikit-image implementation for SLIC is used~\cite{vanderwalt2014scikit}.

\subsection{Sampling}
This work generates samples indicating which segments to perturb using ``only one'', ``all but one'', random, and entropic sampling.
``Only one'' sampling generates all samples where only one segment is perturbed.
``All but one'' sampling generates all samples where all but one segment is perturbed.
Both ``Only one'' and ``All but one'' sampling also generate a sample where no segment is perturbed as a reference value.
Random sampling generates a given number of samples where each segment is perturbed with a probability of $0.5$.
Entropic sampling generates a given number of samples in increasing order of entropy, \ie{} samples with as many or as few segments perturbed as possible.

Entropic sampling is implemented to be similar to the default KernelSHAP sampling behavior~\cite{lundberg2017unified}.
No segments are perturbed in the first sample; all segments are perturbed in the second, followed by all possible combinations of perturbing one segment and all combinations of perturbing all but one segment, followed by combinations of two segments perturbed/unperturbed, and so on until the sample size is reached.

``Only one'' and ``all but one'' sampling create one sample per segment, which makes them computationally efficient.
They are presumably the simplest sampling methods that can be used and have worked well with the Contextual Importance and Utility (CIU) method for detecting bleeding in gastro-enterological~\cite{knapic2021explainable}.
Random and entropic sampling can generate samples consisting of combinations of segments, which can yield many more unique samples.
Using combinations of segments presumably allows for taking into account inter-dependencies between different segments, potentially leading to more informative explanations.

\subsubsection{Sample size}
Random and entropic sampling use sample sizes of 4000/8000, 400, as well as 50.
The 4000/8000 sample size is used to be consistent with the original RISE evaluation.
AlexNet and VGG-16 use 4000 samples, and ResNet models were evaluated with 8000 samples.
``Only one'' and ``all but one'' sampling always generate samples equal to the number of segments plus one, which is 50 samples at most in this work.

\subsection{Perturbing}
This work perturbs images by replacing the perturbed segments with a solid color, which is the most common way of perturbing images.
The color used for perturbing is the mean color of the ImageNet training set, which is the color that will be normalized to 0 in the preprocessing of the Torchvision models used in this work, which is the same as the original RISE implementation.
Like the original RISE implementation, all pixels in the segments are not replaced with exactly the given color and instead are faded between their original color and the normalization mean according to a segment mask that fades closer to the segment border.
This mask is also later reused to attribute per-pixel in addition to the normal per-segment attribution.

Perturbing consists of pixel-wise multiplication between the normalized image and a perturbation mask of values between 0 and 1.
The mask is created by setting all values in the segments to be perturbed to 0 and all others to 1, then the mask is smoothed so that the values closer to the center of each segment are close to 0, and those at the edges and beyond are closer to 1.
Thus, pixels outside the perturbed segments are mostly unchanged, but fade towards the normalization mean as they get closer to the segment centers.
The original implementation achieves this by using bilinear upsampling to scale a $7\times7$ grid of 0s and 1s to the size of the full image, an implementation that is replicated in this work.
An issue with this method is that it relies on having a lower resolution mask to upscale, which excludes using some popular segmentation methods, such as SLIC.
To combat this issue, this work introduces the use of Gaussian filters for mask smoothing.
The Gaussian filter smoothens the mask by assigning each pixel in the mask the average value of all other pixels, weighted by a 2D Gaussian function centered on the pixel.
This work uses a Gaussian filter with $\sigma=10$, which in this use case gives similar smooth masks when compared to bilinear upscaling.

\subsection{Attributing}
This work evaluates five existing attribution methods, CIU~\cite{framling1995extracting}, PDA~\cite{robnik2008explaining}, LIME~\cite{ribeiro2016should}, SHAP~\cite{lundberg2017unified}, and RISE~\cite{petsiuk2018rise}.
In this work, the method names refer only to the actual attribution calculation of each method. 

CIU is presumably the first post-hoc XAI method proposed~\cite{framling1995extracting}, which has also been implemented for explaining image classification in more recent works~\cite{framling2024pyciuimage}.
CIU works by calculating the Contextual Importance ($CI$) of a feature $s$ as $CI_1(s) = \frac{max(Y,Y\backslash_s)-min(Y,Y\backslash_s)}{max(Y\backslash)-min(Y\backslash)}$, where $Y$ is the original output, $Y\backslash_s$ is all the outputs when feature $s$ has been perturbed, and $Y\backslash$ are all outputs.
The CIU implementation for images~\cite{knapic2021explainable} instead calculates the importance of a segment $s$ by perturbing all other segments (``all but one'' sampling).
In this work this is calculated as $CI_2(s) = \frac{max(Y,1-Y\backslash_{\bar{s}})-min(Y,1-Y\backslash_{\bar{s}})}{max(Y\backslash)-min(Y\backslash)}$, where $Y\backslash_{\bar{s}}$ is all the outputs where $s$ is not perturbed.
The Contextual Utility ($CU$) of the feature $s$ is then calculated as $CU(s) = \frac{Y-min(Y\backslash_s)}{max(Y\backslash)-min(Y\backslash)}$ where $Y$ is the original output.
In the case of images, it is usually easiest to use the \textit{contextual influence} as the attribution score for the feature $s$, which in this work is $w_{CIU}(s) = CI(s) \cdot (CU(s) - 0.5)$.
In this work, CIU is only evaluated for the ``only one'' and ``all but one'' sampling methods.
CIU handles interdependencies between pixels and segments through so-called intermediate concepts~\cite{framling2020decision}, which are essentially named coalitions of features, \ie  segments in this case.
Intermediate concepts do not assign value to individual features, and their application to images is ongoing research; as such, it has not been used in this work.

The Prediction Difference Analysis (PDA) method~\cite{robnik2008explaining} works similarly to CIU, but uses average difference instead of maximum difference.
PDA has been adapted to work with images~\cite{wei2018explain}, though both in the original and image implementation, only a single feature is changed at a time (``only one'' sampling).
In this work, PDA has been generalized to work when multiple features are perturbed simultaneously.
The PDA attribution is given by $w_{PDA}(s) = Y - avg(Y\backslash_s)$.

Locally-Interpretable Model-agnostic Explanations (LIME)~\cite{ribeiro2016should} was originally introduced as an umbrella term used to cover any instance where a single prediction is explained by training an interpretable model to mimic the original model's prediction in the neighborhood of the original input.
However, LIME has since been associated with specifically training a linear surrogate model~\cite{ivanovs2021perturbation, schwalbe2023comprehensive} as this is how the method was demonstrated originally.
In this work, LIME is implemented as a linear model $y = b + \sum_{s\in S} w_s\cdot x_s$, where $y$ is the output of the model, $b$ and $w_s$ are the learned bias and weights, and $x_s=0$ if the segment is perturbed and $1$ otherwise.
The attribution of LIME for segment $s$ is the value of $w_s$ after the linear model has been fitted to the input-output pairs using least squares.

Kernel SHAP~\cite{lundberg2017unified} is a modification to LIME such that, under certain assumptions, the weights learned by the linear model will tend towards the Shapley values~\cite{shapley53value}, scoring how the features contribute to the prediction.
This is achieved by scaling the input-output pairs with a kernel function $\pi(X) = \frac{|S|-1}{\binom{|S|}{|X|}|X| (|S|-|X|)}$, where $|s|$ is the number of segments and $|X| = \sum_{x_s\in X} x_s$.
As such the SHAP values can be retrieved by solving $\pi(X) y = b + \sum_{s\in S} w_s\cdot \pi(X)x_s$ using least squares.

The attribution used by RISE~\cite{petsiuk2018rise} is similar to PDA, but instead of using the average decrease when the feature is perturbed, it uses the average prediction when it is not perturbed.
This similarity means that in the case where every possible sample is used, the two methods have equivalent ranks of the segments by influence, which in turn produces equivalent SRG scores.
RISE attribution for a segment is given by $w_{RISE}(s) = avg(Y\backslash_{\bar{s}})$.

In addition to per-segment ($w_s$) attribution, this work also evaluates attribution per-pixel, similar to the original RISE implementation.
Typically, perturbation-based explanations assign the same attribution to all pixels in the same segment ($w_s$).
However, by using the smoothing masks ($M_s$) used for perturbation, the attribution scores of a pixel can instead be set to the average of the attribution of each segment, weighted by how perturbed the pixel was by that segment.
This is formalized as $w_p = \frac{1}{\sum_{s\in S} M_s^p} \sum_{s\in S} w_s\cdot M_s^p$, where $M_s^p$ is the value of pixel $p$ in the perturbation mask of segment $s$.
Note that this calculation means that pixels outside segment $s$, which were slightly perturbed due to the smooth mask, also include that influence in the calculation.
This results in pixels at segment centers getting almost all their attribution score from that segment, while pixels at the borders of segments get a lesser attribution from all the segments they border.

\subsection{Evaluating}
\label{sub:evaluation}
The various pipelines are tested by explaining the predictions of three ImageNet pretrained CNNs on the ImageNet validation set~\cite{deng2009imagenet} and evaluating those explanations using the Symmetric Relevance Gain (SRG) metric~\cite{blucher2024decoupling}.
The three pretrained CNNs are AlexNet~\cite{krizhevsky2012imagenet, krizhevsky2014one}, VGG-16~\cite{simonyan2015very}, and ResNet-50~\cite{he2016deep} using trained parameters from the Torchvision $0.15.2$ framework~\cite{marcel2010torchvision}.
The input to the models is normalized using the average pixel values of ImageNet.

The ImageNet validation set is used as it is the same dataset on which RISE was originally evaluated.
Evaluation is carried out using the first image of each class of the ImageNet validation for a total of 1000 images ($2\%$ of the total validation set).
The limitation to 1000 images is to make the computation demands feasible as over 300 pipelines were evaluated, some requiring 8000 model calls per image.
Despite the limited number of images and some pipelines using the same model calls as they share the first parameters, the evaluation still required close to a billion model calls.
Limited evaluation was performed on the full validation set, and no statistically significant ($p<0.05$) difference could be found compared to using 2\% of the data.
For each image, the top predicted class of each model was explained through segment and pixel attribution using each pipeline.

\begin{figure}[t] 
    \centering
    \includegraphics[width=\columnwidth]{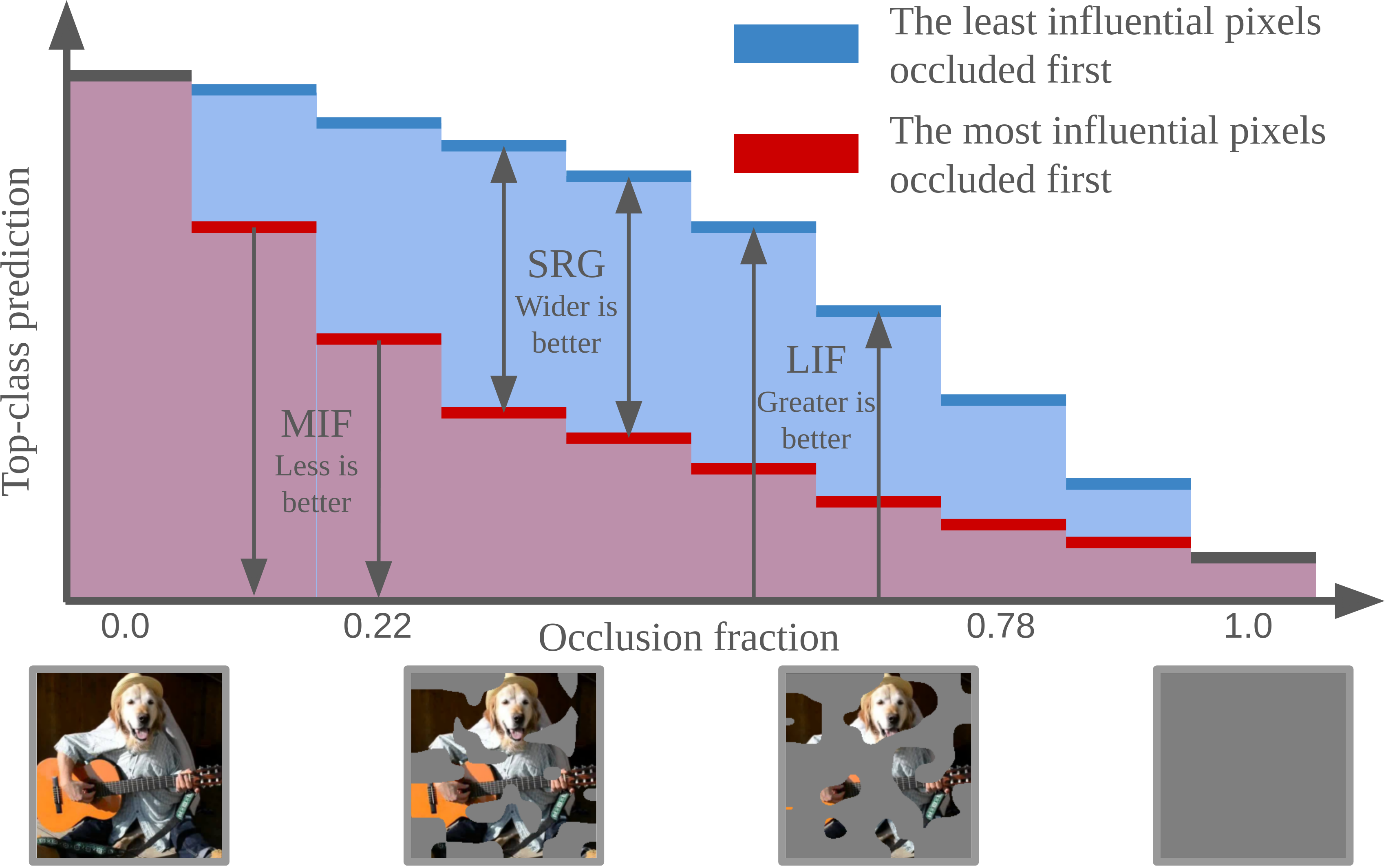}
    \caption{Showcase of how LIF, MIF, and SRG metrics are calculated by steadily occluding the least or most influential pixels of an image and calculating the value of the top class predicted for the original image.}
    \label{fig:srg}
\end{figure}

The SRG metric is used due to the large number of parameter combinations to evaluate, which signifies that human evaluation is not feasible to use for this study.
SRG is measured by increasingly occluding the image and observing how the prediction changes.
By occluding the pixels with the Least Influence First (LIF), the model prediction is expected to be similar until the influential pixels start getting occluded.
Conversely, by occluding the pixels with the Most Influence First (MIF), the model prediction is expected to decrease quickly.
A good explanation should have a large area under the LIF prediction-occlusion curve and a small area under the MIF curve.
LIF and MIF are equivalent to the insertion and deletion metrics used to evaluate RISE originally~\cite{petsiuk2018rise}.
However, LIF and MIF have been shown to be inconsistent at ranking explanations as the occlusion type changes (e.g. random color, mean color, inpainting).
The SRG metric, calculated as the difference between LIF and MIF, is more reliable~\cite{blucher2024decoupling}.
This also eliminates some of the effects of occluded images being out-of-distribution.
The connection between LIF, MIF, and SRG is visualized in Fig.~\ref{fig:srg}.

In this work, SRG is calculated by occluding the image over a total of 10 equal steps (from 0\% occlusion in step 1 to 100\% occlusion in step 10).
The remaining pixels with the lowest or greatest attribution score for the original top-class prediction are occluded in each step for LIF and MIF, respectively.
When there are many pixels with the same attribution, then pixels are chosen in an arbitrary deterministic order.
Occlusion is performed by setting the pixels to a solid gray color, which mirrors the occlusion in the deletion metric used in the work that introduced RISE ~\cite{petsiuk2018rise}.
The average of the original top-class prediction over these 10 images is then recorded as the LIF and MIF scores.
The SRG score is calculated as $LIF-MIF$.



\section{Results and Analysis}
\label{toc:results}

The results consist of the LIF, MIF, and SRG metrics for every attribution pipeline.
As this is too much data to present in this work, it is summarized as the average SRG metric for different parameter combinations.
The tables show for some parameter combinations the SRG in \% averaged over all pipelines with those combinations (\ie{} the average result for all combinations of the unspecified parameters).

The average SRG for pipelines with different sampling methods and sample sizes over the different attribution methods is shown in Table~\ref{tab:attribution_vs_sampling}.
Notably, sampling has a larger impact on performance than the attribution calculation.
Unsurprisingly, increasing sample size yields improved performance.
What is surprising is that random sampling significantly outperforms entropic sampling, even for SHAP, for which entropic sampling is the default sampling method.
This may be caused by the manner in which entropic sampling was implemented in this work as described later.
Furthermore, some attribution methods seem to fail when used together with certain sampling methods and sample sizes.
LIME and SHAP perform poorly when using 50 random samples, likely due to having too little feature-specific information for least squares to have a single useful convergence.
PDA also performs poorly when using entropic sampling at higher sample sizes, the reason for which will be described later in this section.
For non-random sampling with a sample size of 50, all attribution methods have the same performance.
Under these circumstances, CIU, PDA, and RISE mathematically produce the same rankings of segments to which LIME and SHAP consistently converge.

\begin{table}[t] 
    \centering
    \caption{The SRG in \% for all pipelines with different combinations of sampling and attribution methods. The results are averaged for all pipelines with the specified parameters.}
    \label{tab:attribution_vs_sampling}
    \resizebox {\linewidth}{!}{%
    \begin{tabular}{l l | l l l l l}
    \toprule
         & & \multicolumn{5}{c}{Attribution}\\
        Sampling    & Sample size & CIU  & PDA  & LIME & SHAP & RISE\\\hline
        Only one    & $\leq50$    & 13.3 & 13.3 & 13.3 & 13.3 & 13.3\\
        All but one & $\leq50$    & 14.9 & 14.9 & 14.9 & 14.9 & 14.9\\
        Random      & 50          & -    & 16.0 & 6.8  & 6.8  & 16.0\\
        Entropic    & 50          & -    & 13.3 & 13.3 & 13.3 & 13.3\\\hdashline
        Random      & 400         & -    & 22.9 & 24.1 & 22.3 & 22.9\\
        Entropic    & 400         & -    & 9.0  & 15.6 & 17.3 & 15.0\\\hdashline
        Random      & 4000/8000   & -    & 25.6 & 25.8 & 24.0 & 25.6\\
        Entropic    & 4000/8000   & -    & 14.7 & 18.2 & 18.8 & 17.8\\
    \bottomrule
    \end{tabular}
    }
\end{table}

The results of different combinations of segmenting, perturbing, and attribution can be found in Table~\ref{tab:attribution_vs_segmenting}.
Again, the attribution calculation has less impact on performance compared to the other parameters.
Notably, average performance increases when attributing per-pixel for all combinations of segmenting, perturbing, and attribution methods.
The Gaussian filter does not give a significantly different performance for grid segmentation compared to bilinear upsampling.
However, as the Gaussian filter was implemented to mimic bilinear upsampling in this case, the results show that there is no harm in choosing one over the other.
Instead, the benefit of smoothing with a Gaussian filter becomes obvious in the significant performance increase seen with SLIC over grid segmentation.
As SLIC with pixel attribution is not possible with bilinear upsampling, this suggests that Gaussian filters enable the use of better segmentation methods without sacrificing performance on grid segmentation.

\begin{table}[t] 
    \centering
    \caption{The SRG in \% for all pipelines with different combinations of segmenting, perturbing, and attribution methods with either per-segment or per-pixel attribution. The results are averaged for all pipelines with the specified parameters.}
    \label{tab:attribution_vs_segmenting}
    \resizebox {\linewidth}{!}{%
    \begin{tabular}{l l | l l l l l}
    \toprule
         & & \multicolumn{5}{c}{Attribution}\\
        \makecell[l]{Segmenting +\\Perturbing} & \makecell[l]{Attribute\\per} & CIU$^*$  & PDA  & LIME & SHAP & RISE\\\hline
        Grid+bilinear & Segment & 11.7 & 14.9 & 14.9 & 14.5 & 15.9\\
        Grid+bilinear & Pixel   & 14.1 & 16.3 & 16.5 & 16.4 & 17.6\\\hdashline
        Grid+Gaussian & Segment & 11.6 & 14.9 & 15.0 & 14.6 & 15.8\\
        Grid+Gaussian & Pixel   & 14.4 & 16.5 & 16.8 & 16.7 & 17.8\\\hdashline
        SLIC+Gaussian & Segment & 15.7 & 17.1 & 17.4 & 17.5 & 18.0\\
        SLIC+Gaussian & Pixel   & 16.8 & 17.6 & 18.2 & 18.3 & 18.8\\
    \bottomrule
    \multicolumn{7}{l}{$^*$CIU is not evaluated for random or entropic sampling, which have}\\
    \multicolumn{7}{l}{greater average performance.}\\
    \end{tabular}
    }
\end{table}

It is also informative to inspect attribution maps generated for some example images that have been used in the evaluation.
The attribution maps generated are overlaid on the original image to make it easy to see which part of the image has a higher attribution than another.
The pixels are colored in an even spectrum between red and blue according to their rank from most attribution to least.
This is in contrast to how it is commonly done, where each pixel is colored according to the value of its attribution relative to the least and greatest attribution in the image.
The reason for this approach is that it is easier to assess the order in which pixels are occluded in the SRG metric, which does not care for the distribution of attribution, only the order.

\begin{figure*}[t] 
    \centering
    \includegraphics[width=1.5\columnwidth]{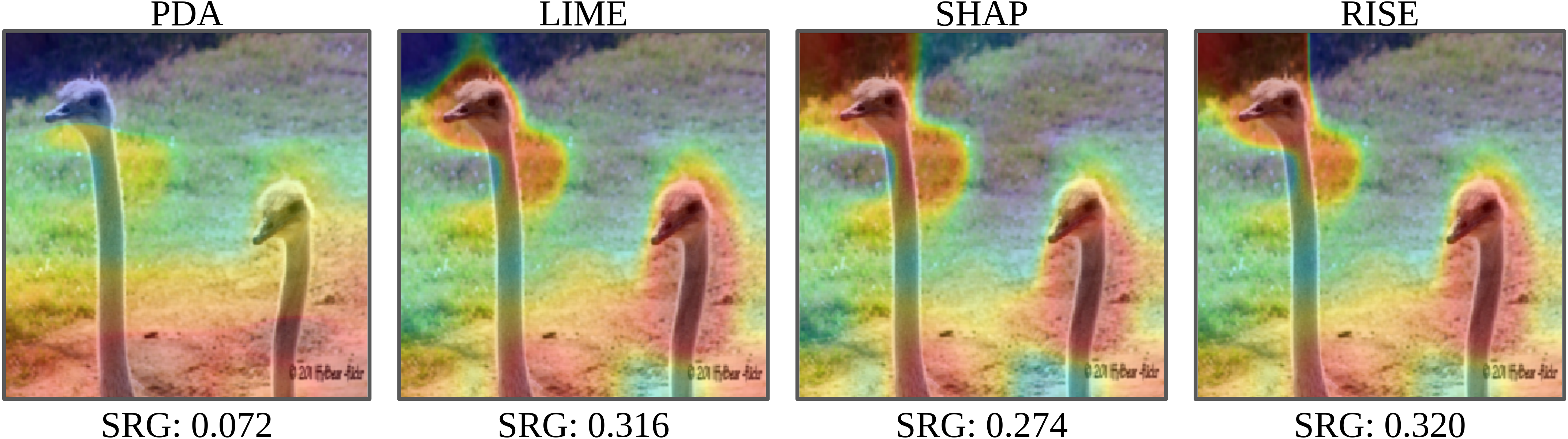}
    \caption{
        Attribution maps per attribution method overlaid an image from the ostrich class explaining the correct prediction by AlexNet.
        The maps are attributed per-pixel and use grid+bilinear segmentiation and entropic sampling with a sample size of 400.
        The pixels are colored in an even spectrum between red and blue according to their rank from most attribution to least.
    }
    \label{fig:ostrich}
\end{figure*}

Fig.~\ref{fig:ostrich} compares the attribution maps generated using the PDA, LIME, SHAP, and RISE attribution methods.
The comparison uses entropic sampling with a sample size of 400 to investigate a scenario where the performance between the methods varies.
Aside from PDA, it is clear that the methods generate very similar attribution maps.
This behavior is consistent between images and sample sizes.
By switching to random sampling, PDA will generate attribution maps that are similar to those of the other methods.
This behavior of PDA, seemingly arbitrarily giving less attribution to the pixels at the top of the image can be used to explain its poor performance with entropic sampling.
The way entropic sampling is performed in this work is the cause of this behavior, as it alternates between all combinations where $x$ features are perturbed and all combinations where all except $x$ features are perturbed, with $x$ increasing from 0.
Since 400, 4000, and 8000 are not neat cutoff points, this systematically leaves some segments being perturbed more often than others.
This affects both PDA and RISE as they respectively use the average output when the feature is and isn't perturbed.
However, since the sampling starts with fewer samples being perturbed, this results in the difference in the number of times each feature is perturbed being higher (ranging from 56 to 98 for sample size 400) than the difference in the number of times it isn't perturbed (ranging from 302 to 344).
This means that PDA is much more severely affected by the imbalance in sampling than RISE is.
To verify this hypothesis, the experiment has been conducted again with the reversed alternation of perturbing and not perturbing features in entropic sampling.
The results of these additional experiments can be found in Table~\ref{tab:reverse_entropic}.
When alternation is reversed, RISE instead suffers from a similar loss of performance as PDA does, while PDA gains a significant boost to performance.
This seems to verify the hypothesis that PDA and RISE are sensitive to unbalanced sampling.

\begin{table}[t] 
    \centering
    \caption{The SRG in \% for PDA and RISE pipelines with entropic sampling where the alternation between perturbing $x$ samples and perturbing all but $x$ samples has been reversed. The results are averaged for all pipelines with the specified parameters.}
    \label{tab:reverse_entropic}
    \resizebox {0.9\linewidth}{!}{%
    \begin{tabular}{l l | l l }
    \toprule
         & & \multicolumn{2}{c}{Attribution}\\
        Sampling                       & Sample size & PDA  & RISE\\\hline
        Entropic (reverse alternation) & 400         & 19.3 & 9.5\\
        Entropic (reverse alternation) & 4000/8000   & 20.5 & 16.4\\
    \bottomrule
    \end{tabular}
    }
\end{table}

\begin{figure*}[t] 
    \centering
    \includegraphics[width=1.5\columnwidth]{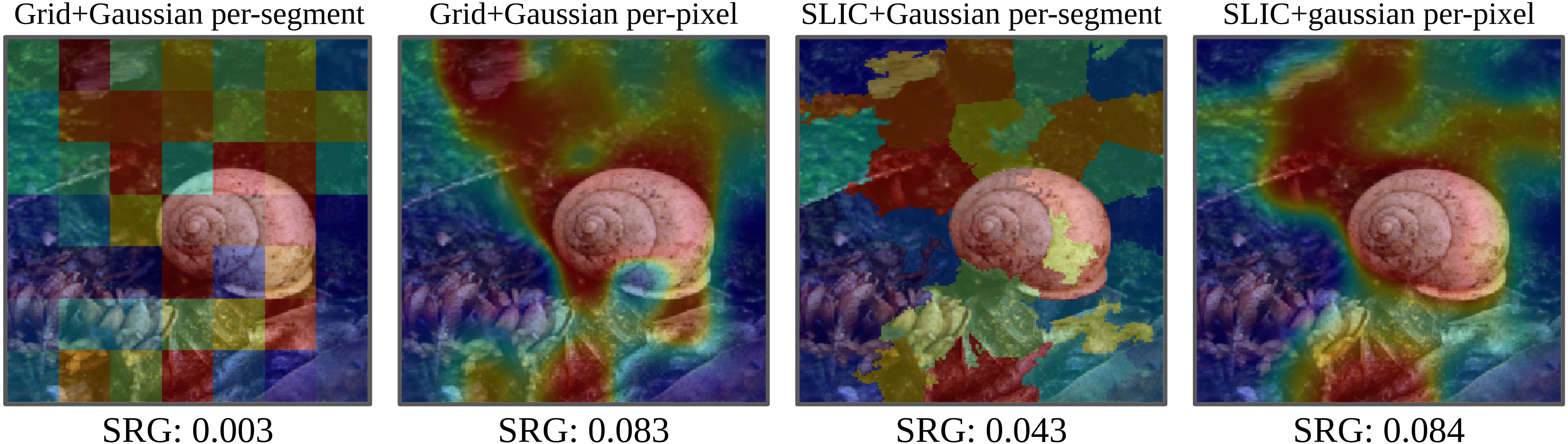}
    \caption{
        Per-segment and per-pixel attribution maps generated using both Grid+Gaussian and SLIC+Gaussian segmentation overlaid an image of a snail, explaining the correct prediction by AlexNet.
        The maps were generated using "only one" sampling and are equivalent across all attribution methods.
        The pixels are colored in an even spectrum between red and blue according to their rank from highest attribution to lowest.
    }
    \label{fig:snail}
\end{figure*}

Fig.~\ref{fig:snail} compares the per-segment and per-pixel attribution maps generated with grid and SLIC segmentation.
The comparison uses Gaussian smoothing for both segmentation methods for a fairer comparison and "only one" sampling to use a case where all attribution methods give the same attribution maps.
The figure highlights one advantage of per-pixel attribution: its ability to smooth the borders between segments means that shapes that do not align with the segments can be better contained in the areas of high or low attribution without a harsh cutoff.
This advantage is more prevalent when using grid segmentation, as the method does not take the shapes of the image into account when segmenting.
SLIC has a similar benefit over grid segmentation as its segments are adapted to the actual features of the image.
That SLIC already fits the features of the image is likely why it gets relatively less improvement from per-pixel attribution compared to grid segmentation.
In the case of this image, the difference in the per-pixel case is essentially negligible.

\section{Discussion}
\label{toc:discussion}

Most works that introduce or compare perturbation-based image explanation methods do not examine the parameters of the explanation pipelines separately.
This leads to a poor understanding of why a method performs well and can lead to questionable claims of some methods being better than others~\cite{velmurugan2023through, miro2024assessing}.
The results of this work show that each parameter of the perturbation-based image explanations pipeline significantly impacts performance.

From the results, it is clear that using SLIC as opposed to grid segmenting improves performance significantly.
Per-pixel attribution, which requires smooth masks, also improves performance compared to per-segment attribution.
Finally, using SLIC with per-pixel attribution further improved performance.
This shows that using Gaussian filters or other alternatives to bilinear smoothing is promising, as it allows the use of better-performing segmentation methods.
The choice of attribution calculation has a noticeably smaller impact on performance when compared to the choice of segmentation method or the difference between per-segment and per-pixel attribution.
The largest impact on performance came from issues that PDA and RISE attribution encounter when sampling is biased towards certain features.
It is likely that using more careful sampling methods would eliminate these differences, making the choice of attribution method seemingly arbitrary when it comes to occlusion metrics.
This is supported by visualizations of the attribution maps generated for the images, which are similar.
When it comes to feature attribution for images, it is arguable that performance is not the most relevant criterion for choosing an attribution method.
Instead, it is useful to consider other factors when choosing an attribution method.
CIU, PDA, and RISE are all likely easier to explain to a lay user than LIME and SHAP, which is an advantage.
CIU's ability to differentiate between importance and influence is specific to that attribution method.
On the other hand, CIU, in its current form, cannot assign attribution to single features when using sampling methods other than "only one" and "all but one", which limits its performance on occlusion metrics in high sampling situations.


It is clear that an increased sample size does improve performance.
However, as model calls dominate the computational demands of the other parts of the pipelines, the total computation of each pipeline scales directly with sample size.
With ever-increasing computational demands by newer DNN models, even a low sample size, let alone thousands of samples, may be unrealistic to presume for an explanation of a single decision.
In some cases, such as medical diagnosis prediction, the need for and the value of explanations are likely high enough that it is worth increasing computational demands by factors of thousands, but for many other cases, low-computation explanations are needed.
Developing perturbation-based methods that can give good explanations with a low sample size is, therefore, a promising future direction.
One approach to this could be by using a limited number of samples initially and using the results from those samples to determine the next set of samples.
Such an approach would essentially make the sampling, perturbing, and predicting steps of the pipelines recurrent.

The evaluation in this work relied on the explanation methods being separable into different parameters that could be combined in various ways.
This is not always the case, even if the method otherwise produces sound explanations.
For example, the original RISE implementation shifts the perturbation masks by some pixels so as not to center the same pixels every time.
This approach works with the RISE attribution calculation since it can directly assign influence to pixels.
However, this is not feasible for other methods, as such shifting has not been used in this evaluation.
Additionally, the use of SRG and similar metrics requires attribution scores for individual features.
For example, the CIU method can use intermediate concepts to explain combinations of features, which results in explanations for the combinations rather than for the individual features.

This work shows how different parameters affect performance as measured by SRG.
SRG measures how well the explanation reflects the actual impact of the attributed regions on the model.
While ensuring that attribution maps explain the model correctly is helpful, it is not known whether it correlates with usefulness to humans.
Existing evaluations on whether proxy metrics correspond to usefulness for humans are limited and require further investigation~\cite{nguyen2021effectiveness}.
It is also questionable whether proxy metrics can be used to compare different types of attribution maps at all.
Attribution maps that indicate large regions, that provide heatmaps, or that show important edges and textures are likely to be interpreted differently by humans in a way not reflected by most proxy metrics.

Finally, all the steps of the perturbation-based image attribution pipeline need to be considered when implementing an explanation system.
Future research should be careful when comparing different pipelines with each other, not to assign performance increases to some individual part of a pipeline unless the other parameters are the same.
Similar pipelines likely exist in other domains beyond images, which would likely also benefit from further investigation.

\bibliography{biblio}
\bibliographystyle{IEEEtranN}

\end{document}